\documentclass{article}

\usepackage{arxiv}

\usepackage{algorithm}
\usepackage{algorithmic}

\usepackage{amsmath}
\usepackage{txfonts}

\usepackage[utf8]{inputenc} 
\usepackage[T1]{fontenc}    
\usepackage{hyperref}       
\usepackage{url}            
\usepackage{booktabs}       
\usepackage{amsfonts}       
\usepackage{nicefrac}       
\usepackage{microtype}      
\usepackage{cleveref}       
\usepackage{lipsum}         
\usepackage{graphicx}
\usepackage{natbib}
\usepackage{doi}

\title{Attention Is Not All You Need Anymore}

\date{}

\newif\ifuniqueAffiliation
\uniqueAffiliationtrue

\ifuniqueAffiliation 
\author{ {\hspace{1mm}Zhe~Chen} \\
	School of Computer Science and Engineering\\
	Northeastern University \\
          Shenyang, Liaoning, China\\
	\texttt{ml\_iot@163.com; chenzhe@mail.neu.edu.cn} \\
}
\else
\usepackage{authblk}

\newbox{\orcid}\sbox{\orcid}{\includegraphics[scale=0.06]{orcid.pdf}} 
\fi

\renewcommand{\headeright}{Zhe Chen}
\renewcommand{\undertitle}{}
\renewcommand{\shorttitle}{Attention Is Not All You Need Anymore}

\begin{document}
\maketitle

\begin{abstract}
In recent years, the popular Transformer architecture has achieved great success in many application areas, including natural language processing and computer vision. Many existing works aim to reduce the computational and memory complexity of the self-attention mechanism in the Transformer by trading off performance. However, performance is key for the continuing success of the Transformer.
In this paper, a family of drop-in replacements for the self-attention mechanism in the Transformer, called the Extractors, is proposed. Four types of the Extractors, namely the super high-performance Extractor (SHE), the higher-performance Extractor (HE), the worthwhile Extractor (WE), and the minimalist Extractor (ME), are proposed as examples. Experimental results show that replacing the self-attention mechanism with the SHE evidently improves the performance of the Transformer, whereas the simplified versions of the SHE, i.e., the HE, the WE, and the ME, perform close to or better than the self-attention mechanism with less computational and memory complexity.
Furthermore, the proposed Extractors have the potential or are able to run faster than the self-attention mechanism since their critical paths of computation are much shorter.
Additionally, the sequence prediction problem in the context of text generation is formulated using variable-length discrete-time Markov chains, and the Transformer is reviewed based on our understanding.

\end{abstract}


\section{Introduction}

It has been over six years since the introduction of the widely adopted Transformer architecture~\cite{van17}. While initially proposed for sequence transduction tasks, the Transformer has become the de-facto model architecture for a broad range of natural language processing tasks in recent years~\cite{kim2023big} and has been widely applied in many other areas including computer vision~\cite{khan2022} and speech processing~\cite{latif2023transformers}.

As of July 2023, the standard Transformer with the self-attention mechanism still prevails, especially in large language models (LLM)~\cite{touvron2023llama}, due to its excellent parallelizability and capacity~\cite{zhao2023survey}.

The key to the success of the Transformer lies in the self-attention mechanism~\cite{dowdell2019attention}, a type of the attention mechanism first introduced for machine translation in~\cite{ahdanau14}.

However, a notable limitation of the Transformer is its quadratic computation and memory complexity in relation to sequence length. This limitation arises from the self-attention mechanism~\cite{ren21}, posing a challenge for applying the Transformer in scenarios involving long input sequences.

Therefore, many variants, including ``efficient Transformers'' or ``$x$-formers'', have been proposed to address this issue in the past few years~\cite{tay23}. In general, these variants leverage pre-determined patterns or learnable patterns, low-rank approximations, kernels, downsampling, or sparsification to reduce the computation and memory complexity in the self-attention mechanism. As a result, they generally underperform the vanilla self-attention mechanism~\cite{dong2023survey}.

Truly, the Transformer with the self-attention mechanism is a remarkable piece of artwork. It is quite challenging to surpass it in terms of performance. In order to improve its performance, we need to comprehensively understand the problem we face and every bit of the Transformer and consider replacing the self-attention mechanism, which is its  Achilles' heel. After all, self-attention is not an indispensable ingredient of the Transformer~\cite{liu21, mehta23, tolstikhin21}. It is time to move beyond the self-attention mechanism.

In this paper, we first formulate the sequence prediction problem (e.g., predicting the next token based on the preceding tokens) in text generation using variable-length discrete-time Markov chains. Then, the Transformer is reviewed based on our understanding. Furthermore, we propose a family of drop-in replacements for the self-attention mechanism, called the Extractors. The Transformers equipped with the proposed Extractors are verified in a text generation scenario. 
Experimental results show that with the Extractors replacing the self-attention mechanism, the Transformers either perform much better with more computational and memory complexity or perform close to or better with less computational and memory complexity, confirming that the self-attention mechanism is indeed not all we need anymore.

Our contributions are summarized as follows.
\begin{itemize}
  \item  We employ variable-length discrete-time Markov chains to formulate the sequence prediction problem in text generation.
  \item  We review the Transformer based on our own understanding.
  \item  We propose a family of sublayers called the Extractors to replace the self-attention sublayer in the Transformer in a drop-in fashion.
  \item  We evaluate the performance of the Transformers equipped with the Extractors in a text generation setting using free English children's books as the dataset. 
  \item  We estimate the computational and memory complexities of both the self-attention sublayer and the proposed Extractor sublayers.
\end{itemize}

\section{Related Work}

Several solutions have been proposed to replace the self-attention mechanism in a drop-in fashion.

In the area of computer vision, a multi-layer perceptron (MLP) is employed to replace the self-attention in~\cite{tolstikhin21} with linear computational complexity in sequence length, achieving slightly inferior performance to other models.

In~\cite{tay2022pretrained}, the self-attention is replaced with convolutional blocks. The performance is not as good as that of the standard Transformer in tasks that require modeling the relationship between two or more sequences.

In~\cite{thorp22}, the Fourier transform is employed to replace the self-attention mechanism. The performance is close to that of the standard Transformer and the training speed is faster.

SPADE (state space augmented transformer) uses a state space model (SSM) in the bottom layer and efficient local attention mechanisms in other layers~\cite{zuo2022efficient}. The performance of SPADE is close to that of the standard Transformer, and it has a linear time and space computational complexity.

A subquadratic drop-in replacement for attention named Hyena is proposed in~\cite{poli2023hyena}. Hyena alternatively applies convolutions in the time domain and in the frequency domain. It achieves comparable performance to the Transformer with a 20\% reduction in training compute required at a sequence length of 2k.

\section{Problem Formulation}

In this paper, we use text generation as an example task for the Transformer. In text generation, additional tokens are generated based on the given tokens. In this task, a token refers to a piece of text, e.g., a word. All the valid tokens make up the vocabulary. To facilitate computation, each token in the vocabulary is associated with an index ranging from $0$ to $u-1$, where $u$ is the size of the vocabulary. 

The fundamental problem in text generation is sequence prediction, i.e., predicting the value of the next token index $s_{t+1}$ based on the values of a sequence of given token indices $(s_1,s_2,\cdots,s_t)$, where $s_1,s_2,\cdots,s_{t+1}\in\{0,1,\cdots,u-1\}$ and $t$ is the length of the given input sequence. Since token indices are countable and can take any numbers in $\{0,1,\cdots,u-1\}$, discrete random variables $S_1,S_2,\cdots,S_{t+1}$ can be used to associate with the token indices at time $1$, time $2$, $...$, time $t+1$, respectively. 

To predict the value of the next token index $s_{t+1}$ more accurately, we may try to make good use of the values of the token indices in the given sequence. Since the length of the sequence is not fixed in general in text generation, a reasonable way is to view the sequence as a variable-length discrete-time Markov chain~\cite{buhl99, chen10}, where the value of the next token index only depends on the values of the token indices in the given sequence with a maximum sequence length. 

In this way, we define the conditional probability of taking the value of the next token index $s_{t+1}$, which is the time-homogeneous state-transition probability of the Markov chain, as follows. 
\begin{equation} 
  \label{eq_p}
  P\left(S_{t+1}=s_{t+1}\middle| S_t=s_t,S_{t-1}=s_{t-1},\cdots,S_1=s_1\right)
\end{equation} 
where $P(\cdot)$ denotes probability function. Let $l$ denote the length of the context window in text generation, which is also the highest order of the variable-length discrete-time Markov chain. When $t>l$, according to the Markov property we have
\begin{equation}
\begin{aligned}
  \label{eq_pl}
  P\left(S_{t+1}=s_{t+1}\middle| S_t=s_t,S_{t-1}=s_{t-1},\cdots,S_1=s_1\right)= P\left(S_{t+1}=s_{t+1}\middle| S_t=s_t,S_{t-1}=s_{t-1},\cdots,S_{t-l+1}=s_{t-l+1}\right)
\end{aligned}
\end{equation} 

Thus, the sequence prediction problem we are facing turns into the problem of predicting the probability of taking value $s_{t+1}$ given a sequence of values $(s_1,s_2,\cdots,s_t)$.

In text generation, if we have the probabilities $P\left(S_{t+1}=s_{t+1}\middle| S_t=s_t,S_{t-1}=s_{t-1},\cdots,S_1=s_1\right)$ for all $s_{t+1}$ in ${\{0,1,\cdots,u-1\}}$, strategies such as top-$p$ sampling and top-$k$ sampling can be employed to choose a value for $S_{t+1}$.

But how can we get these probabilities? In machine learning, we try building a model to predict or infer these probabilities by taking the sequence $(s_1,s_2,\cdots,s_t)$ as the input to the model. This is where the Transformer is applied in text generation.

The next question is how to train the model. Since we have no idea what these probabilities should look like before we train the model, a strategy we employ here is simply maximizing the predicted probability $\hat{P}\left(S_{t+1}=s_{t+1}\middle| S_t=s_t,S_{t-1}=s_{t-1},\cdots,S_1=s_1\right)$ that the model outputs given an observed sequence $(s_1,s_2,\cdots,s_t,s_{t+1})$. With this strategy, the loss function for training the model given a training sequence of length $l$ can be derived as below. This is the same as what we get if we alternatively regard sequence prediction as a multi-class classification task.
\begin{equation}
  \label{eq_loss}
  L\left(\boldsymbol{W}\right)=-\frac{1}{l}\sum_{t=1}^{l}\ln{\left(\hat{P}\left(S_{t+1}=s_{t+1}\middle| S_t=s_t,\cdots,S_1=s_1\right)\right)}
\end{equation} 
where $L\left(\cdot\right)$ denotes the loss function and $\boldsymbol{W}$ generally refers to all the trainable model parameters. To keep the problem simple, in the rest of this paper, we assume $t\le l$.

\section{Understanding The Transformer} \label{transformer}

The Transformer provides us with an effective way, but of course not necessarily the only effective way, to build the aforementioned machine learning model. A distinctive feature of the Transformer is that it keeps its internal ``dimension'' constant across all layers, which makes the Transformer look like a kind of transformation in general.

\begin{figure}[!t]
  \centering
  \includegraphics[width=0.45\columnwidth]{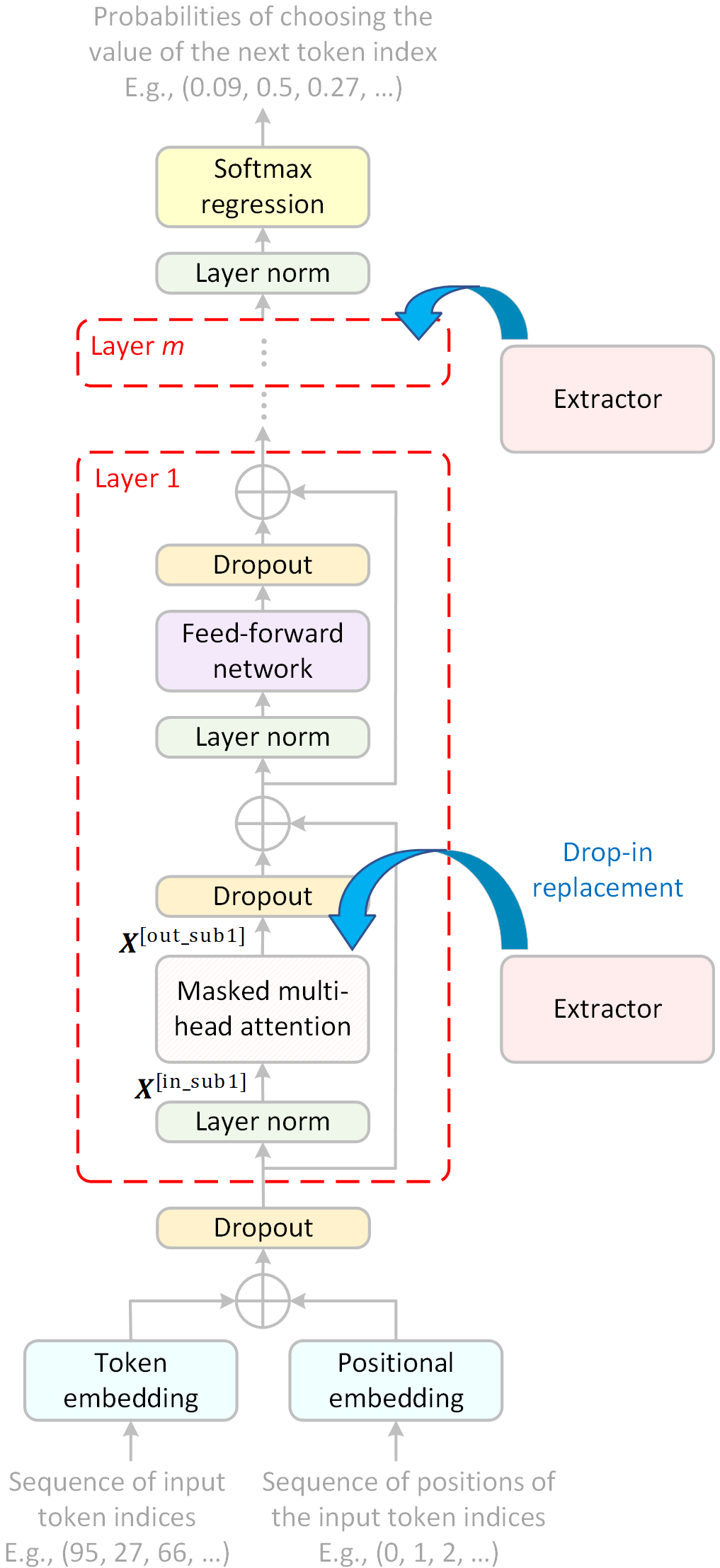} 
  \caption{The Transformer.}
  \label{fig_trans}
\end{figure}
As shown in Fig.~\ref{fig_trans}, the input to the Transformer is a sequence of token indices denoted by $\boldsymbol{s}={(s_1,s_2,\cdots,s_t)}^{\mathrm{T}}$, where $\boldsymbol{s}\in\mathbb{Z}^{t\times1}$ and ${(\cdot)}^{\mathrm T}$ denotes transpose. The sequence of token indices $\boldsymbol{s}$ is virtually converted into one-hot encodings $\boldsymbol{X}^{[\mathrm{in\_tok}]}$ before it goes through a linear transformation called ``embedding'', as shown in Eq.~(\ref{eq_emb_tok}), where $\boldsymbol{X}^{[\mathrm{in\_tok}]}={({\boldsymbol x}_i^{[\mathrm{in\_tok}]} )}_{1 \le i \le t}^{\mathrm{T}}$, $\boldsymbol{X}^{[\mathrm{in\_tok}]} \in \mathbb{Z}^{t\times u}$, ${\boldsymbol x}_i^{[\mathrm{in\_tok}]}=([s_i=j-1])_{1 \le j \le u}$, ${\boldsymbol x}_i^{[\mathrm{in\_tok}]} \in \mathbb{Z}^{1 \times u}$, and $\left[\cdot\right]$ is the Iverson bracket, $\left[s_i=j-1\right]=1$ if $s_i=j-1$ otherwise $\left[s_i=j-1\right]=0$. 
\begin{equation} 
  \label{eq_emb_tok}
  \boldsymbol{X}^{[\mathrm{emb\_tok}]}=\boldsymbol{X}^{[\mathrm{in\_tok}]} \boldsymbol{W}^{[\mathrm{emb\_tok}]} 
\end{equation} 
where $\boldsymbol{X}^{[\mathrm{emb\_tok}]} \in \mathbb{R}^{t \times d}$, $\boldsymbol{W}^{\left[\mathrm{emb\_tok}\right]}$ is a weight matrix, $\boldsymbol{W}^{\left[\mathrm{emb\_tok}\right]}\in\mathbb{R}^{u\times d}$, and $d$ is the internal ``dimension'' of the Transformer.

In this paper, we use learned positional embeddings, since their performance is nearly identical to that of the sinusoidal version~\cite{van17}. Then we have
\begin{equation} 
  \label{eq_emb_pos}
  \boldsymbol{X}^{[\mathrm{emb\_pos}]}=\boldsymbol{X}^{[\mathrm{in\_pos}]} \boldsymbol{W}^{[\mathrm{emb\_pos}]} 
\end{equation} 
where $\boldsymbol{X}^{[\mathrm{emb\_pos}]} \in \mathbb{R}^{t \times d}$, $\boldsymbol{X}^{[\mathrm{in\_pos}]}={({\boldsymbol x}_i^{[\mathrm{in\_pos}]} )}_{1 \le i \le t}^{\mathrm{T}}$, $\boldsymbol{X}^{[\mathrm{in\_pos}]} \in \mathbb{Z}^{t \times l}$, ${\boldsymbol x}_i^{[\mathrm{in\_pos}]}=([i=j])_{1 \le j \le l}$, ${\boldsymbol x}_i^{[\mathrm{in\_pos}]} \in \mathbb{Z}^{1 \times l}$, $\boldsymbol{W}^{[\mathrm{emb\_pos}]}$  is a weight matrix, and $\boldsymbol{W}^{[\mathrm{emb\_pos}]} \in \mathbb{R}^{l \times d}$. Note that according to~\cite{van17}, embedding weight matrices such as $\boldsymbol{W}^{[\mathrm{emb\_tok}]}$ and $\boldsymbol{W}^{[\mathrm{emb\_pos}]}$ are mulitiplied by $\sqrt{d}$.

Then, the two embedding matrices are added together, and the elements of their sum matrix are randomly set to zeros with a probability of $p$ during training, which is called ``dropout''. Dropout is a regularization technique that enhances the performance of the model during inference, albeit at the expense of decreased model ``capacity''. As depicted in Eq.~(\ref{eq_dropout}), dropout is applied to each element $x_{i,j}^{[\mathrm{in\_drop}]}$ of its input matrix $\boldsymbol{X}^{[\mathrm{in\_drop}]}$, where $ \boldsymbol{X}^{[\mathrm{in\_drop}]}=({x_{i,j}^{[\mathrm{in\_drop}]}})_{1 \le i \le t,1 \le j \le d}$ and $\boldsymbol{X}^{[\mathrm{in\_drop}]} \in \mathbb{R}^{t \times d}$.
\begin{equation} 
  \label{eq_dropout}
  x_{i,j}^{[\mathrm{out\_drop}]}=\left\{\begin{matrix}\frac{1}{1-p}x_{i,j}^{[\mathrm{in\_drop}]} & \mathrm{with~probability}~1-p \\ 0 & \mathrm{with~probability}~p \\\end{matrix}\right.
\end{equation} 
where $x_{i,j}^{[\mathrm{out\_drop}]}$  is the element of its output matrix $\boldsymbol{X}^{[\mathrm{out\_drop}]}$ and $\boldsymbol{X}^{[\mathrm{out\_drop}]} \in \mathbb{R}^{t \times d}$. 
To simplify equations, we denote dropout as the ``function'' $\mathrm{dropout}(\cdot)$. Then, the input to the first layer of the Transformer $\boldsymbol{X}^{[\mathrm{in\_lay1}]}$ can be written as
\begin{equation} 
  \label{eq_in_layer}
  \boldsymbol{X}^{[\mathrm{in\_lay1}]}=\mathrm{dropout}(\boldsymbol{X}^{[\mathrm{emb\_tok}]} + \boldsymbol{X}^{[\mathrm{emb\_pos}]}) 
\end{equation} 
where $\boldsymbol{X}^{[\mathrm{in\_lay1}]} \in \mathbb{R}^{t \times d}$.

As shown in Fig.~\ref{fig_trans}, there are two sublayers in a layer of the Transformer. In the standard Transformer, the first sublayer is the multi-head self-attention sublayer, and the second sublayer is a feed-forward network (FFN). The inputs to the sublayers first go through layer normalizations, which are called pre-layer normalizations, and dropouts are applied to the outputs of the sublayers. Pre-layer normalization is employed in this paper since Transformers with pre-layer normalizations can be trained without the warm-up stage~\cite{xiong20b}. There are residual connections between the inputs of the layer normalizations and the outputs of the dropouts. Both pre-layer normalizations and residual connections are beneficial for easing the training process. 
Layer normalization is applied to each row vector  $\boldsymbol{x}_i^{[\mathrm{in\_layernorm}]}$ of its input matrix $\boldsymbol{X}^{[\mathrm{in\_layernorm}]}$, where $\boldsymbol{X}^{[\mathrm{in\_layernorm}]}=({\boldsymbol{x}_i^{[\mathrm{in\_layernorm}]}})^{\mathrm{T}}_{1 \le i \le t}$ and $\boldsymbol{X}^{[\mathrm{in\_layernorm}]} \in \mathbb{R}^{t \times d}$, as shown in Eq.~(\ref{eq_layernorm})~\cite{ba2016layer}.
\begin{equation} 
  \label{eq_layernorm}
  \boldsymbol{x}_i^{[\mathrm{out\_layernrom}]}=\boldsymbol{g}^{[\mathrm{layernorm}]} \circ \frac{\boldsymbol{x}_i^{[\mathrm{in\_layernorm}]}-{\mu}_i}{{\sigma}_i}+\boldsymbol{b}^{[\mathrm{layernorm}]}
\end{equation} 
where $\boldsymbol{g}^{[\mathrm{layernorm}]}$ is the gain vector and $\boldsymbol{b}^{[\mathrm{layernorm}]}$ is the bias vector, $\boldsymbol{g}^{[\mathrm{layernorm}]}, \boldsymbol{b}^{[\mathrm{layernorm}]} \in \mathbb{R}^{1 \times d}$, ${\mu}_i$ and ${\sigma}_i$ are the mean and standard deviation of the elements of $\boldsymbol{x}_i^{[\mathrm{in\_layernorm}]}$, respectively, $\boldsymbol{x}_i^{[\mathrm{out\_layernrom}]}$  is the $i$-th row vector of the output matrix $\boldsymbol{X}^{[\mathrm{out\_layernorm}]}$, $\boldsymbol{X}^{[\mathrm{out\_layernorm}]} \in \mathbb{R}^{t \times d}$, $i=1, 2, \cdots, t$, and $\circ$ denotes element-wise product or Hadamard product. To simplify equations, we denote layer normalization as the ``function'' $\mathrm{layernorm}(\cdot)$. Then, the input to the first sublayer $\boldsymbol{X}^{[\mathrm{in\_sub1}]}$ can be obtained following Eq.~(\ref{eq_in_sub1}), where $\boldsymbol{X}^{[\mathrm{in\_sub1}]} \in \mathbb{R}^{t \times d}$.
\begin{equation} 
  \label{eq_in_sub1}
  \boldsymbol{X}^{[\mathrm{in\_sub1}]}=\mathrm{layernorm}(\boldsymbol{X}^{[\mathrm{in\_lay1}]})
\end{equation} 

\begin{figure}[!t]
  \centering
  \includegraphics[width=0.8\columnwidth]{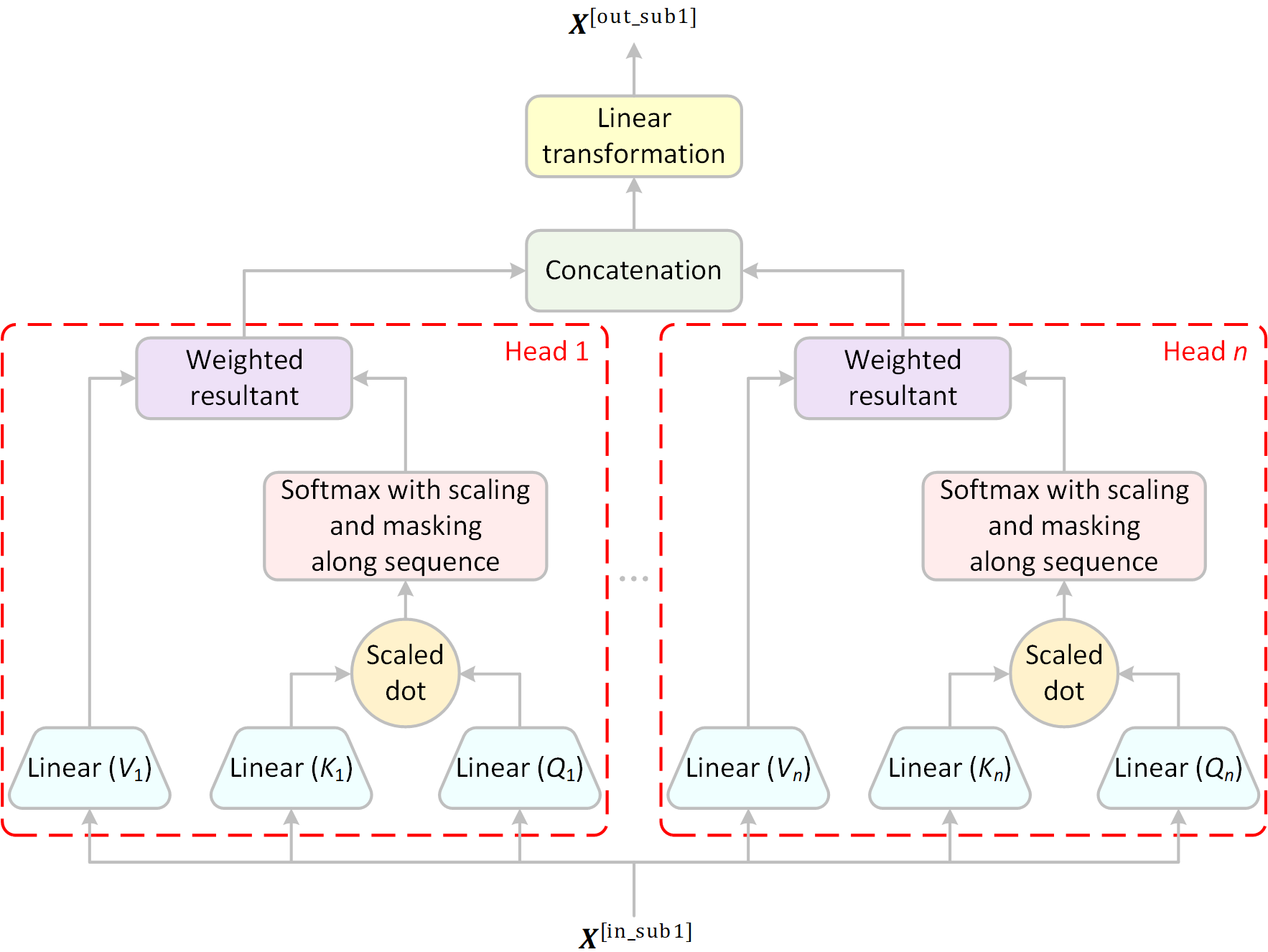} 
  \caption{The multi-head self-attention sublayer.}
  \label{fig_att}
\end{figure}
Fig. 2 depicts the multi-head self-attention sublayer. The input $\boldsymbol{X}^{[\mathrm{in\_sub1}]}$ is first transformed into $n$ sets of query/key/value matrices, namely $\boldsymbol{Q}_j$, $\boldsymbol{K}_j$, and $\boldsymbol{V}_j$, where $j=1, 2, \cdots, n$ and $n$ is the number of ``heads''.
\begin{equation} 
  \label{eq_qj}
  \boldsymbol{Q}_j=\boldsymbol{X}^{\mathrm{[in\_sub1}]} \boldsymbol{W}_j^{[\mathrm Q]}
\end{equation} 
\begin{equation} 
  \label{eq_kj}
  \boldsymbol{K}_j=\boldsymbol{X}^{\mathrm{[in\_sub1}]} \boldsymbol{W}_j^{[\mathrm K]}
\end{equation} 
\begin{equation} 
  \label{eq_vj}
  \boldsymbol{V}_j=\boldsymbol{X}^{\mathrm{[in\_sub1}]} \boldsymbol{W}_j^{[\mathrm V]}
\end{equation} 
where $\boldsymbol{Q}_j, \boldsymbol{K}_j, \boldsymbol{V}_j\in\mathbb{R}^{t\times h}$, $\boldsymbol{W}_j^{[\mathrm Q]}$, $\boldsymbol{W}_j^{[\mathrm K]}$, $\boldsymbol{W}_j^{[\mathrm V]}$ are weight matrices, $\boldsymbol{W}_j^{[\mathrm Q]}, \boldsymbol{W}_j^{[\mathrm K]}, \boldsymbol{W}_j^{[\mathrm V]} \in \mathbb{R}^{d\times h}$, and $h=\frac{d}{n}$. Then, scaled dot products are calculated using both query matrices and key matrices, as shown in Eq.~(\ref{eq_zj}).
\begin{equation} 
  \label{eq_zj}
  \boldsymbol{Z}_j^{[\mathrm{att}]}=\frac{\boldsymbol{Q}_j \boldsymbol{K}_j^{\mathrm T}}{\sqrt h}
\end{equation} 
where $\boldsymbol{Z}_j^{[\mathrm{att}]} \in \mathbb{R}^{t\times t}$ and $j=1,2,\cdots,n$. After that, softmax is applied to the scaled dot products to output normalized weights $\boldsymbol{A}_j$, where $\boldsymbol{A}_j=\left(a_{j,i,k}\right)_{1\le i\le t,\ 1\le k\le t}$, $\boldsymbol{A}_j\in\mathbb{R}^{t\times t}$, $j=1,2,\cdots,n$, and 
\begin{equation} 
  \label{eq_aijk}
  a_{j,i,k}=\left\{\begin{matrix}\frac{\mathrm{e}^{z_{j,i,k}}}{\sum_{f=1}^{i}\mathrm{e}^{z_{j,i,f}}}&k=1,2,\cdots,i\\0&k=i+1,\cdots,t\\\end{matrix}\right.
\end{equation} 

Note that masking is also applied at the same time in Eq.~(\ref{eq_aijk}). Then, weighted resultants are computed for each ``head'' using these normalized weights, and the resultants are simply put together to produce a matrix, which is called concatenation.
\begin{equation} 
  \label{eq_rav}
  \boldsymbol{R}_j=\boldsymbol{A}_j\boldsymbol{V}_j
\end{equation} 
\begin{equation} 
  \label{eq_out_con}
  \boldsymbol{X}^{[\mathrm{out\_con}]}=(\boldsymbol{R}_1,\boldsymbol{R}_2,\cdots,\boldsymbol{R}_n )
\end{equation} 
where $\boldsymbol{X}^{[\mathrm{out\_con}]} \in \mathbb{R}^{t \times d}$, $\boldsymbol{R}_j\in\mathbb{R}^{t\times h}$, and $j=1,2,\cdots,n$. The output of the concatenation $\boldsymbol{X}^{[\mathrm{out\_con}]}$ goes through a linear transformation, which concludes the multi-head self-attention sublayer, as shown in Eq.~(\ref{eq_out_mod}).
\begin{equation} 
  \label{eq_out_mod}
  \boldsymbol{X}^{[\mathrm{out\_sub1}]}=\boldsymbol{X}^{[\mathrm{out\_con}]} \boldsymbol{W}^{[\mathrm{out\_att}]} 
\end{equation} 
where $\boldsymbol{X}^{[\mathrm{out\_sub1}]}$ is the output of the multi-head self-attention sublayer, $\boldsymbol{X}^{[\mathrm{out\_sub1}]}\in\mathbb{R}^{t \times d}$, $\boldsymbol{W}^{[\mathrm{out\_att}]}$ is a weight matrix, $\boldsymbol{W}^{[\mathrm{out\_att}]}\in\mathbb{R}^{d \times d}$.
According to Fig.~\ref{fig_trans}, the input to the second sublayer, or the FFN sublayer, $\boldsymbol{X}^{[\mathrm{in\_sub2}]}$ can be obtained following Eq.~(\ref{eq_in_sub2}), where $\boldsymbol{X}^{[\mathrm{in\_sub2}]} \in \mathbb{R}^{t \times d}$.
\begin{equation} 
  \label{eq_in_sub2}
  \boldsymbol{X}^{[\mathrm{in\_sub2}]}=\mathrm{layernorm}(\mathrm{dropout}(\boldsymbol{X}^{[\mathrm{out\_sub1}]}) + \boldsymbol{X}^{[\mathrm{in\_lay1}]})
\end{equation} 
And the output of the FFN sublayer $\boldsymbol{X}^{[\mathrm{out\_sub2}]}$ is produced according to Eq.~(\ref{eq_out_sub2}).
\begin{equation} 
\begin{aligned}
  \label{eq_out_sub2}
  \boldsymbol{X}^{[\mathrm{out\_sub2}]}=\mathrm{ReLU}(\boldsymbol{X}^{[\mathrm{in\_sub2}]} \boldsymbol{W}^{[\mathrm{sub2\_1}]}  + \boldsymbol{b}^{[\mathrm{sub2\_1}]}) \boldsymbol{W}^{[\mathrm{sub2\_2}]} + \boldsymbol{b}^{[\mathrm{sub2\_2}]}
\end{aligned}
\end{equation} 
where $\mathrm{ReLU}(\cdot)$ denotes the rectified linear unit (ReLU) activation function, $\boldsymbol{W}^{\left[\mathrm{sub2\_1}\right]}$ and $\boldsymbol{W}^{\left[\mathrm{sub2\_2}\right]}$ are weight matrices, $\boldsymbol{W}^{[\mathrm{sub2\_1}]} \in \mathbb{R}^{d \times c}$, $\boldsymbol{W}^{[\mathrm{sub2\_2}]} \in \mathbb{R}^{c \times d}$, $\boldsymbol{b}^{\left[\mathrm{sub2\_1}\right]}$ and $\boldsymbol{b}^{\left[\mathrm{sub2\_2}\right]}$ are bias vectors, $\boldsymbol{b}^{[\mathrm{sub2\_1}]} \in \mathbb{R}^{1 \times c}$, $\boldsymbol{b}^{[\mathrm{sub2\_2}]} \in \mathbb{R}^{1 \times d}$, and $c$ is the number of the nodes in the hidden layer of the FFN. 
The output of the first layer of the Transformer $\boldsymbol{X}^{[\mathrm{out\_lay1}]}$ can be computed following Eq.~\ref{eq_out_lay1}, where $\boldsymbol{X}^{[\mathrm{out\_lay1}]} \in \mathbb{R}^{t \times d}$.
\begin{equation} 
  \label{eq_out_lay1}
  \boldsymbol{X}^{[\mathrm{out\_lay1}]}=\mathrm{dropout}(\boldsymbol{X}^{[\mathrm{out\_sub2}]}) + \mathrm{dropout}(\boldsymbol{X}^{[\mathrm{out\_sub1}]}) + \boldsymbol{X}^{[\mathrm{in\_lay1}]}
\end{equation} 

There are $m$ layers stacked together in the Transformer as depicted in Fig.~\ref{fig_trans}. Eq.~(\ref{eq_in_sub1}) to Eq.~(\ref{eq_out_lay1}) are repeated until we obtain the output of the $m$-th layer $\boldsymbol{X}^{[\mathrm{out\_lay}m]}$. Since sequence prediction in text generation can also be viewed as a multi-class classification task, the last component in the Transformer is actually a softmax regression, along with pre-layer normalization. 
The output of the softmax regression or the output of the Transformer $\hat{\boldsymbol{Y}}$ is derived as shown in Eq.~(\ref{eq_out_y}).
\begin{equation} 
  \label{eq_out_y}
  \hat{\boldsymbol{Y}}=\mathrm{softmax}\left(\mathrm{layernorm}(\boldsymbol{X}^{\left[\mathrm{out\_lay}m\right]})\boldsymbol{W}^{\left[\mathrm{soft}\right]}+\boldsymbol{b}^{\left[\mathrm{soft}\right]}\right)
\end{equation} 
where $\mathrm{softmax}(\cdot)$ denotes the softmax activation function that takes the rows of its input matrix, $\boldsymbol{W}^{\left[\mathrm{soft}\right]}$ is a weight matrix, $\boldsymbol{W}^{[\mathrm{soft}]} \in \mathbb{R}^{d \times u}$, $\boldsymbol{b}^{\left[\mathrm{soft}\right]}$ is a bias vector, $\boldsymbol{b}^{[\mathrm{soft}]} \in \mathbb{R}^{1 \times u}$, $\hat{\boldsymbol{Y}}=\left({\hat{y}}_{i,j}\right)_{1\le i\le t,\ 1\le j\le u}$, and $\hat{\boldsymbol{Y}}\in\mathbb{R}^{t\times u}$. Then we have
\begin{equation} 
  \label{eq_y_hat}
  \hat{P}\left(S_{i+1}=j-1\middle| S_i=s_i,S_{i-1}=s_{i-1},\cdots,S_1=s_1\right)={\hat{y}}_{i,j}
\end{equation} 
where $i=1, 2,\cdots, t$ and $j=1, 2,\cdots, u$.

\section{The Proposed Extractors}

In this section, we propose a family of drop-in replacements named the Extractors to replace the self-attention mechanism in the Transformer. A distinctive feature of the Extractors is that the weights are assigned in accordance with the reverse order of their input sequence. As examples, we specify four types of the Extractors in this section.

In our view, the FFN sublayer somehow implements a ``lookup table'' or a ``mapper'' and thus ``memorizes'' the relationship between its input and its output, which is beneficial for memorizing the state transitions in the Markov chain. Therefore, if the length or the order of the Markov chain is constant, we may solve the sequence prediction problem just using the FFN sublayer. However, the length of the Markov chain we encounter in text generation is variable, which motivates us to address the variable-length issue in the Markov chain. A reasonable approach to tackle this issue is to convert the ``variable-length'' problem into a ``constant-length'' problem, leading us to the idea of designing the Extractors.

The primal idea of the Extractors is somewhat straightforward. Firstly, we extract unified features from the entire variable-length input sequence as a constant-length ``index'' for ``looking up'' the ``table''. Then, we adjust the unified features according to the length of the sequence since the ``index'' is expected to reflect the length of the sequence. 
That’s it. With the constant-length ``index'' (as well as the input of the layer), we ``look up'' the ``table'', which is implemented by the succedent FFN sublayer, and output the corresponding ``memorized content'' that ``documents'' in which direction the state of the Markov chain should transition.

\subsection{The Super High-Performance Extractor}

\begin{figure}[!t]
  \centering
  \includegraphics[width=0.8\columnwidth]{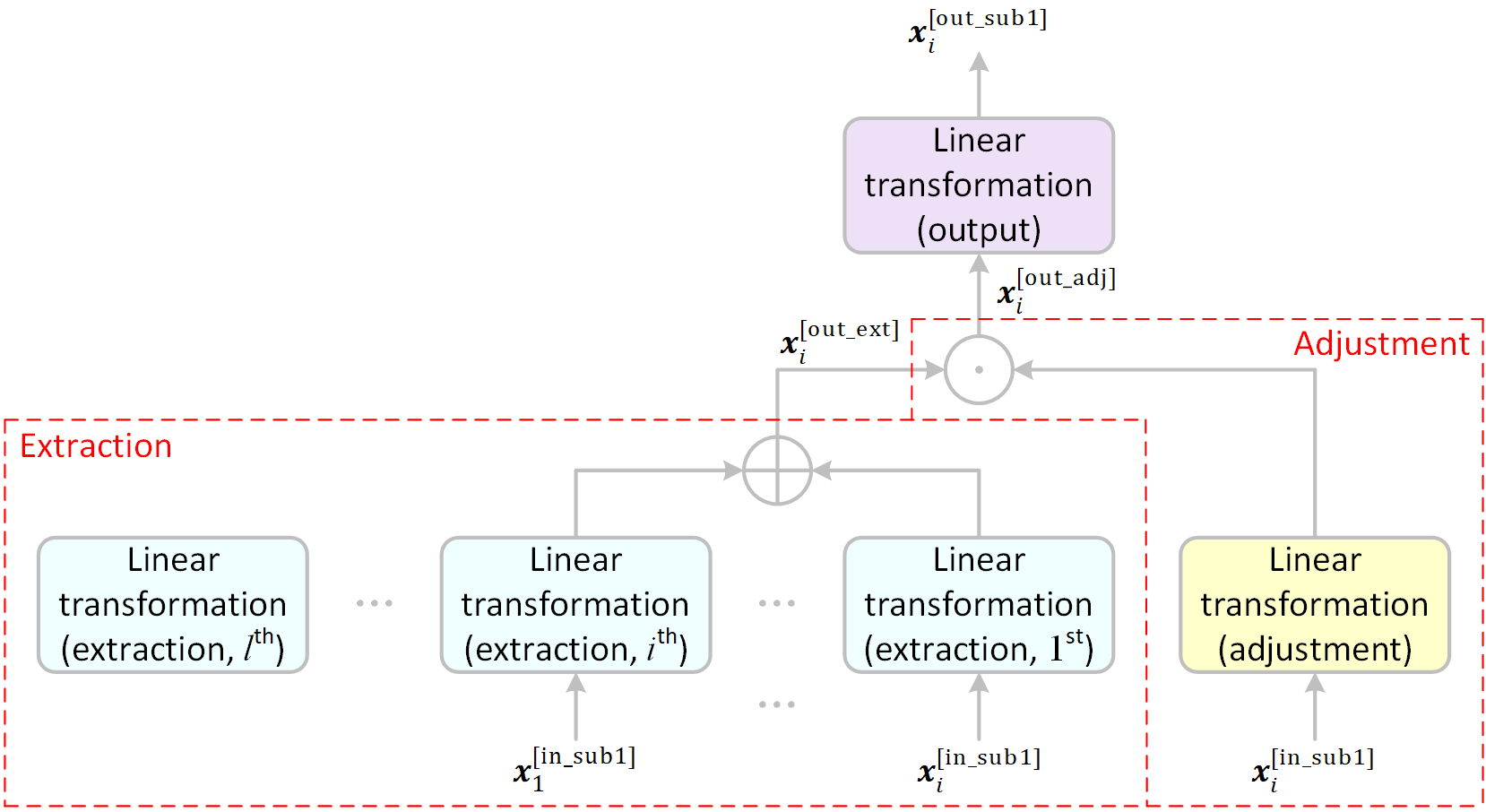} 
  \caption{The proposed SHE.}
  \label{fig_ext_xl}
\end{figure}
Fig.~\ref{fig_ext_xl} illustrates a type of the proposed Extractor called the super high-performance Extractor (SHE). The linear transformations marked in light blue on the left aim to extract unified features from their inputs. The total number of these linear transformation blocks is $l$. Since we assume $t\le l$, only $t$ linear transformation blocks are in use at a certain time. The design of this ``extraction'' part is inspired by both recurrent neural networks (RNNs) and finite impulse response (FIR) filters. 

Given an input matrix to the Extractor sublayer $\boldsymbol{X}^{\left[\mathrm{in\_sub1}\right]}$, where $\boldsymbol{X}^{[\mathrm{in\_sub1}]}\in\mathbb{R}^{t \times d}$, $\boldsymbol{X}^{\left[\mathrm{in\_sub1}\right]}={(\boldsymbol{x}_j^{\left[\mathrm{in\_sub1}\right]})}^{\mathrm T}_{1\le j\le t}$, and $\boldsymbol{x}_j^{\left[\mathrm{in\_sub1}\right]}\in\mathbb{R}^{1\times d}$, the output of the ``extraction'' part $\boldsymbol{X}^{\left[\mathrm{out\_ext}\right]}$ is computed following Eq.~(\ref{eq_out_ext}).
\begin{equation} 
  \label{eq_out_ext}
  \boldsymbol{x}_i^{[\mathrm{out\_ext}]}=\sum_{j=1}^i \boldsymbol{x}_j^{[\mathrm{in\_sub1}]} \boldsymbol{W}_{i-j+1}^{[\mathrm{ext}]}
\end{equation} 
where $\boldsymbol{X}^{\left[\mathrm{out\_ext}\right]}={(\boldsymbol{x}_i^{\left[\mathrm{out\_ext}\right]})}^{\mathrm T}_{1\le i\le t}$, $\boldsymbol{X}^{[\mathrm{out\_ext}]}\in\mathbb{R}^{t \times d}$, $\boldsymbol{x}_i^{\left[\mathrm{out\_ext}\right]}\in\mathbb{R}^{1\times d}$, $\boldsymbol{W}_1^{[\mathrm{ext}]}, \boldsymbol{W}_2^{[\mathrm{ext}]},\cdots,\boldsymbol{W}_l^{[\mathrm{ext}]}$ are weight matrices, $\boldsymbol{W}_1^{[\mathrm{ext}]}, \boldsymbol{W}_2^{[\mathrm{ext}]},\cdots,\boldsymbol{W}_l^{[\mathrm{ext}]}\in\mathbb{R}^{d\times d}$, and $i=1,2,\cdots,t$.

The linear transformation marked in light yellow on the right in Fig.~\ref{fig_ext_xl} intends to output element-wise adjustments for the ``extraction'' part. This ``adjustment'' part takes $\boldsymbol{X}^{\left[\mathrm{in\_sub1}\right]}$ as its input as well. The idea behind this choice is that the latest input vector $\boldsymbol{x}_t^{[\mathrm{in\_sub1}]}$ in each input sequence ${(\boldsymbol{x}_j^{\left[\mathrm{in\_sub1}\right]})}^{\mathrm T}_{1\le j\le t}$ contains the positional information that reflects the length of the sequence and contains valuable information for predicting the next state of the Markov chain. 
The output of the ``adjustment'' part  $\boldsymbol{X}^{[\mathrm{out\_adj}]}$ is derived as follows.
\begin{equation} 
  \label{eq_out_adj}
  \boldsymbol{X}^{[\mathrm{out\_adj}]}=(\boldsymbol{X}^{[\mathrm{in\_sub1}]} \boldsymbol{W}^{[\mathrm{adj}]}) \circ \boldsymbol{X}^{[\mathrm{out\_ext}]}
\end{equation} 
where $\boldsymbol{X}^{[\mathrm{out\_adj}]}\in\mathbb{R}^{t \times d}$, $\boldsymbol{X}^{\left[\mathrm{out\_adj}\right]}={(\boldsymbol{x}_i^{\left[\mathrm{out\_adj}\right]})}^{\mathrm T}_{1\le i\le t}$, $\boldsymbol{x}_i^{\left[\mathrm{out\_adj}\right]}\in\mathbb{R}^{1\times d}$, $\boldsymbol{W}^{[\mathrm{adj}]}$ is a weight matrix, and $\boldsymbol{W}^{[\mathrm{adj}]}\in\mathbb{R}^{d \times d}$.

We can take $\boldsymbol{X}^{[\mathrm{out\_adj}]}$ as the output of the Extractor sublayer, for the following linear transformation shown in Eq.~(\ref{eq_out_sub1_ext}) is optional, as it is not a critical component to the Extractor. 
\begin{equation} 
  \label{eq_out_sub1_ext}
  \boldsymbol{X}^{[\mathrm{out\_sub1}]}=\boldsymbol{X}^{[\mathrm{out\_adj}]} \boldsymbol{W}^{[\mathrm{out}]}
\end{equation} 
where $\boldsymbol{X}^{[\mathrm{out\_sub1}]}$ is the output of the Extractor sublayer, $\boldsymbol{X}^{[\mathrm{out\_sub1}]} \in\mathbb{R}^{t \times d}$, $\boldsymbol{X}^{\left[\mathrm{out\_sub1}\right]}={(\boldsymbol{x}_i^{\left[\mathrm{out\_sub1}\right]})}^{\mathrm T}_{1\le i\le t}$, $\boldsymbol{x}_i^{\left[\mathrm{out\_sub1}\right]}\in\mathbb{R}^{1\times d}$, $\boldsymbol{W}^{\left[\mathrm{out}\right]}$ is a weight matrix, and $\boldsymbol{W}^{[\mathrm{out}]} \in\mathbb{R}^{d \times d}$.

\subsection{The Worthwhile Extractor}

The ``extraction'' part of the SHE virtually uses shared-weight fully-connected neural networks to ``extract'' constant-length ``indices'' and may require significant computational and memory resources. A simplified version of the SHE sublayer called the worthwhile Extractor (WE) is proposed in this subsection. 

\begin{figure}[!t]
  \centering
  \includegraphics[width=0.6\columnwidth]{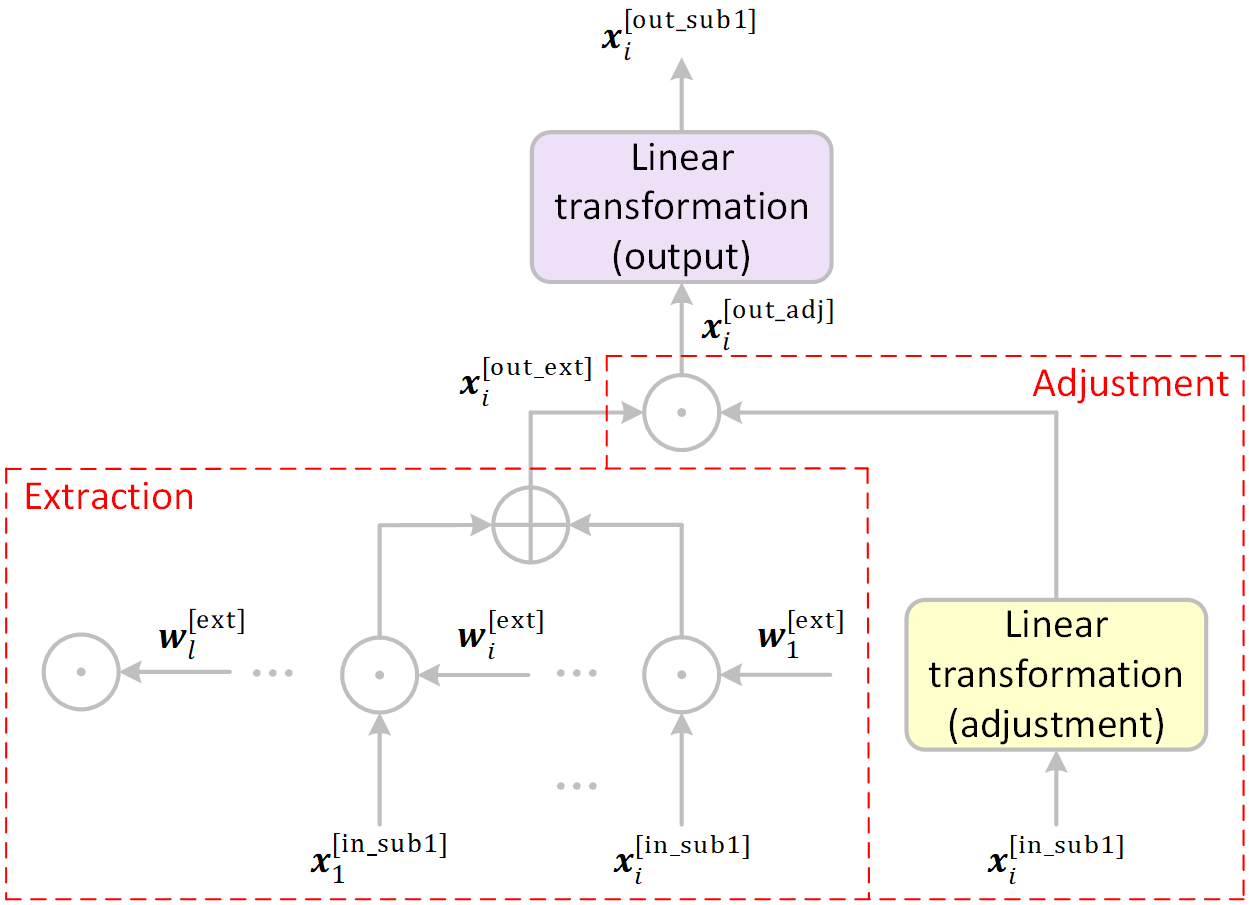} 
  \caption{The proposed WE.}
  \label{fig_ext_m}
\end{figure}
As shown in Fig.~\ref{fig_ext_m}, linear transformations in the ``extraction'' part of the SHE sublayer are replaced by element-wise products. Thus, Eq.~(\ref{eq_out_ext}) is reduced to Eq.~(\ref{eq_out_ext_m}).
\begin{equation} 
  \label{eq_out_ext_m}
  \boldsymbol{x}_i^{[\mathrm{out\_ext}]}=\sum_{j=1}^i \boldsymbol{x}_j^{[\mathrm{in\_sub1}]}  \circ \boldsymbol{w}_{i-j+1}^{[\mathrm{ext}]}
\end{equation} 
where $\boldsymbol{w}_1^{[\mathrm{ext}]}, \boldsymbol{w}_2^{[\mathrm{ext}]},\cdots,\boldsymbol{w}_l^{[\mathrm{ext}]}$ are weight vectors, $\boldsymbol{w}_1^{[\mathrm{ext}]}, \boldsymbol{w}_2^{[\mathrm{ext}]},\cdots,\boldsymbol{w}_l^{[\mathrm{ext}]}\in\mathbb{R}^{1 \times d}$, and $i=1,2,\cdots,t$.

\subsection{The Higher-Performance Extractor}

\begin{figure}[!t]
  \centering
  \includegraphics[width=0.6\columnwidth]{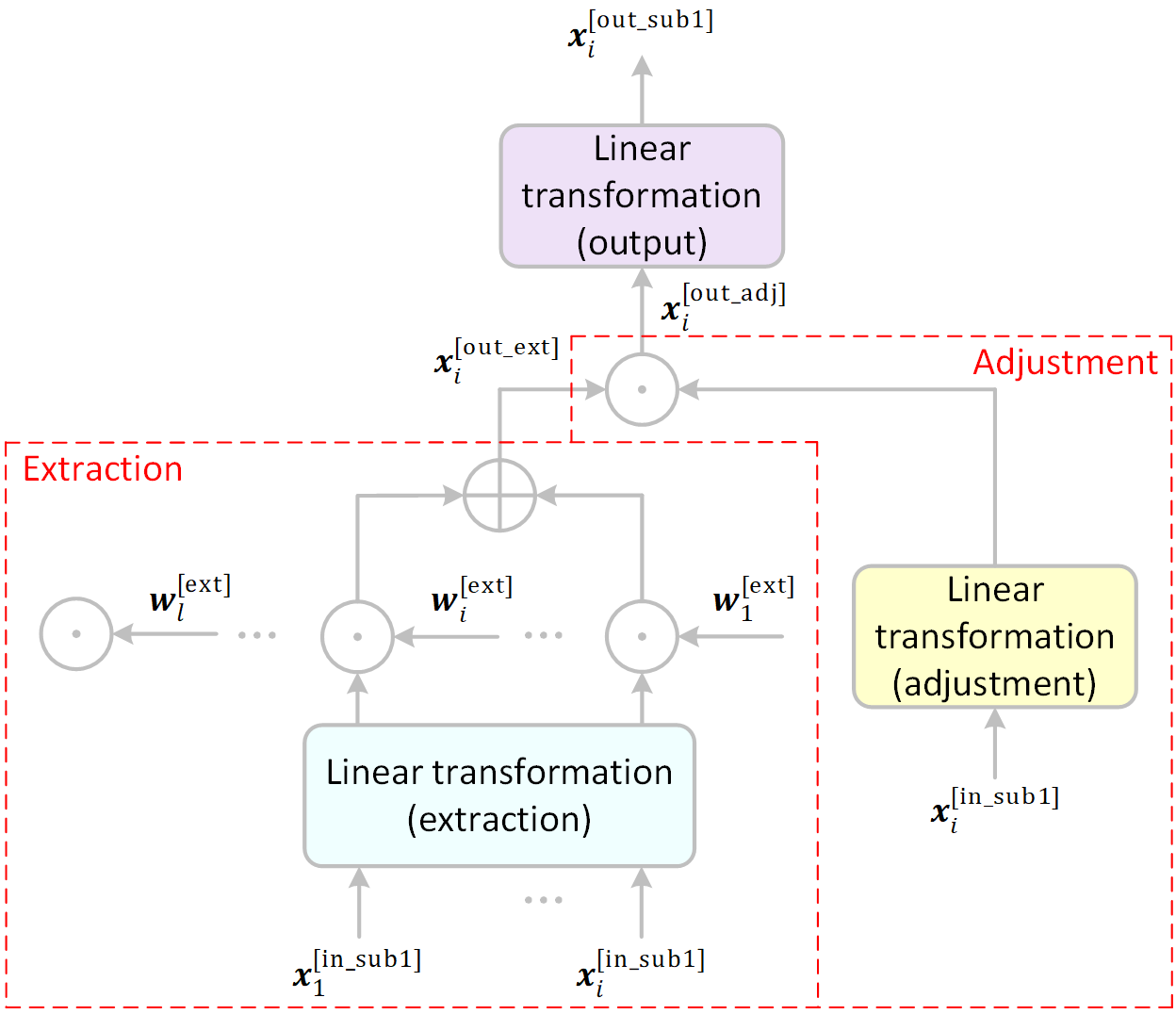} 
  \caption{The proposed HE.}
  \label{fig_ext_l}
\end{figure}
To further enhance the performance of the WE, we incorporate a shared linear transformation in the ``extraction'' part to enable the WE to approximate the SHE,  as depicted in Fig.~\ref{fig_ext_l}. We refer to this type of the Extractor as the higher-performance Extractor (HE). In this case, Eq.~(\ref{eq_out_ext_m}) is replaced by Eq.~(\ref{eq_out_ext_shared}) and Eq.~(\ref{eq_out_ext_l}). 
\begin{equation} 
  \label{eq_out_ext_shared}
  \boldsymbol{X}^{[\mathrm{ext}]}=\boldsymbol{X}^{[\mathrm{in\_sub1}]} \boldsymbol{W}^{[\mathrm{in\_ext}]}
\end{equation} 
where $\boldsymbol{X}^{[\mathrm{ext}]}$ is the output of the linear transformation, $\boldsymbol{X}^{[\mathrm{ext}]} \in\mathbb{R}^{t \times d}$, $\boldsymbol{X}^{\left[\mathrm{ext}\right]}={(\boldsymbol{x}_i^{\left[\mathrm{ext}\right]})}^{\mathrm T}_{1\le i\le t}$, $\boldsymbol{x}_i^{\left[\mathrm{ext}\right]}\in\mathbb{R}^{1\times d}$, $\boldsymbol{W}^{\left[\mathrm{in\_ext}\right]}$ is a weight matrix, and $\boldsymbol{W}^{[\mathrm{in\_ext}]} \in\mathbb{R}^{d \times d}$.
\begin{equation} 
  \label{eq_out_ext_l}
  \boldsymbol{x}_i^{[\mathrm{out\_ext}]}=\sum_{j=1}^i \boldsymbol{x}_j^{[\mathrm{ext}]}  \circ \boldsymbol{w}_{i-j+1}^{[\mathrm{ext}]}
\end{equation} 

\subsection{The Minimalist Extractor}

The computational and memory complexity of the aforementioned Extractors can be further reduced. In this subsection, we propose a simple Extractor called the minimalist Extractor (ME).

\begin{figure}[!t]
  \centering
  \includegraphics[width=0.42\columnwidth]{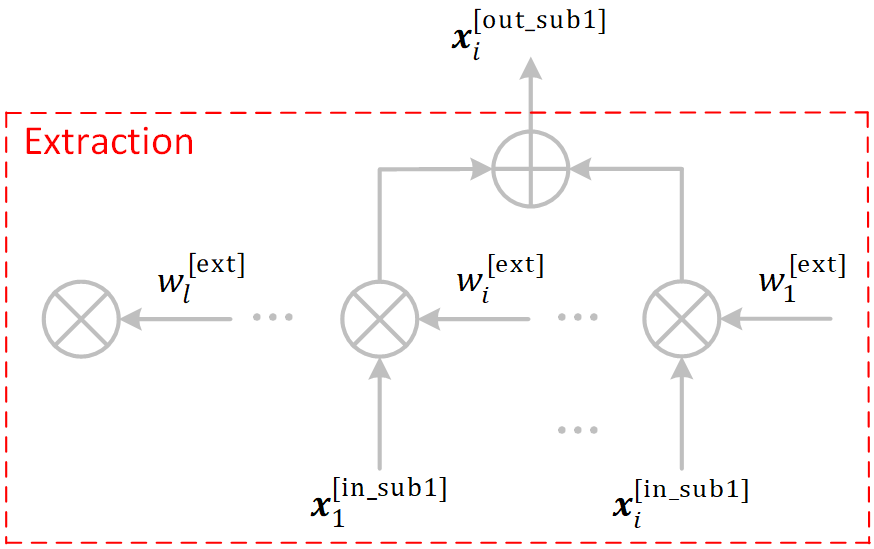} 
  \caption{The proposed ME.}
  \label{fig_ext_s}
\end{figure}
As illustrated in Fig.~\ref{fig_ext_s}, the ME only contains the ``extraction'' part, while the weight vectors in the WE or the HE are further reduced to scalars. Eq.~(\ref{eq_out_ext_s}) describes what the ME does.
\begin{equation} 
  \label{eq_out_ext_s}
  \boldsymbol{x}_i^{[\mathrm{out\_sub1}]}=\sum_{j=1}^i \boldsymbol{x}_j^{[\mathrm{in\_sub1}]}  w_{i-j+1}^{[\mathrm{ext}]}
\end{equation} 
where $w_1^{[\mathrm{ext}]}, w_2^{[\mathrm{ext}]},\cdots, w_l^{[\mathrm{ext}]}$ are weight scalars, $w_1^{[\mathrm{ext}]}, w_2^{[\mathrm{ext}]},\cdots, w_l^{[\mathrm{ext}]}\in\mathbb{R}$, and $i=1,2,\cdots,t$.

Last but not least, the Extractors are not limited to the above four types. Many variations of the four example types of the Extractors can also be used to replace the self-attention mechanism. They are trade-offs between performance and computational and memory complexity. For example, the Extractor that consists of merely the ``extraction'' part of the HE.

\section{Experiments}

In order to evaluate the performance of the proposed Extractors, we replace the multi-head self-attention sublayer in the Transformer introduced in Section~\ref{transformer} with the Extractor sublayers and employ text generation as an example task. It should be noted that Transformers with the Extractor sublayers can also be applied to other tasks. 

Experiments are conducted on an NVIDIA GeForce RTX 4050 GPU (graphics processing unit) with a memory size of 6GB, due to a very limited budget. For this reason, the scale of the models in our experiments is constrained. Therefore, the vocabulary size used in commonplace LLMs may be too large for conducting the experiments. 
To address this issue, we build a small-scale dataset for training the models in the experiments using only popular free English children’s books, since unfortunately we did not find such a publicly available dataset. 
Specifically, the dataset is composed of the top 100 books on English children's literature available at gutenberg.org, a library of free ebooks. The raw text of the books is further tokenized using the Hugging Face BPE (byte-pair encoding) tokenizer with a vocabulary size of 5000, resulting in a total of 8.4M tokens.

We use training cost (i.e., the average training loss over a batch) as the evaluation matric since training cost equals perplexity in this task. Perplexity measures how well a probability model predicts. The lower the perplexity, the better the model predicts.

The models are implemented and trained using the PyTorch 2.0 framework. The biases of the models are initialized with zeros, while the weights are initialized randomly following a normal distribution with a mean of zero and a standard deviation of 0.01. The hyperparameters and settings for the experiments are listed in Table~\ref{table_param}.
\begin{table*}[t]
\centering
\begin{tabular}{l|l}
    \toprule
    Hyperparameter or setting & Value \\
   \midrule
    Size of the vocabulary ($u$) & 5000 \\
    Length of the context window ($l$) & 128 \\
    ``Dimension'' of the model ($d$) & 128 \\
    Number of the nodes in the hidden layer of the FFN ($c$) & 512 \\
    Number of layers ($m$) & 18 \\
    Batch size & 64 \\
    Number of batches for training & 60000 (0.457 epochs) \\
    Optimizer & AdamW ($\gamma=0.001$, $\beta_1=0.9$, $\beta_2=0.999$) \\
    Learning rate & 0.001 \\
    Dropout rate ($p$) & 0.1 \\
    \bottomrule
\end{tabular}
\caption{Hyperparameters and settings for the experiments.}
\label{table_param}
\end{table*}

In order to fairly evaluate the performance of the Transformer with the proposed Extractors and the Transformer with the self-attention mechanism, we train all the models with the same hyperparameters, settings, and training data. By ``the same training data'', we mean that not only all the batches in the training are the same, but also the orders of the batches are the same. This can be implemented by setting the random seed a couple of times.

\begin{figure}[!t]
  \centering
  \includegraphics[width=0.7\columnwidth]{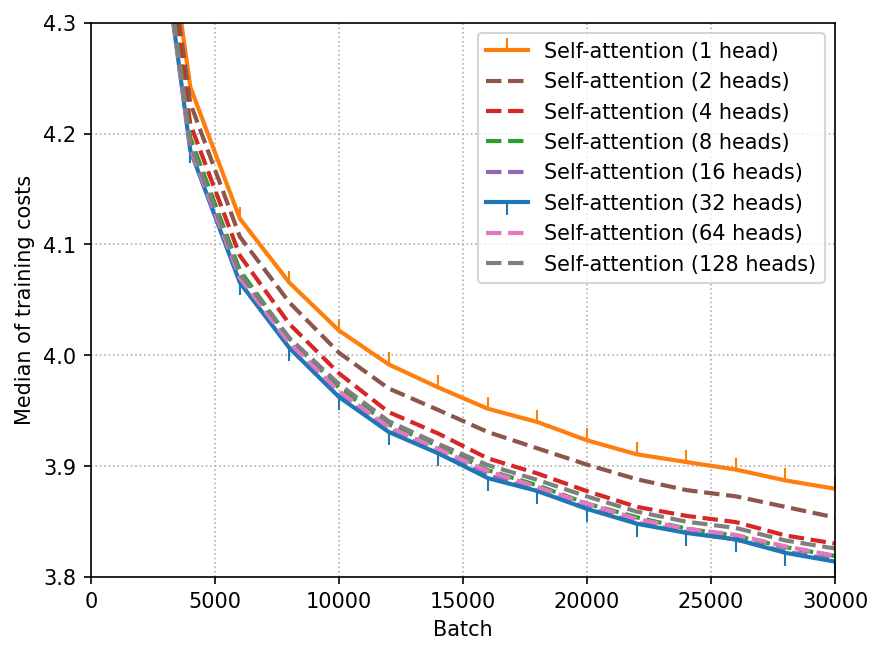} 
  \caption{Medians of the training costs of the models with the self-attention sublayers.}
  \label{fig_att_comp}
\end{figure}
Since the number of heads in the self-attention sublayer slightly affects the performance, eight Transformer models with different numbers of heads (1, 2, 4, 8,16, 32, 64, and 128, respectively) are trained for comparison purposes.  Fig.~\ref{fig_att_comp} shows the medians of the training costs for every non-overlapping 2000 batches. From the figure, we can see that the model with the 1-head self-attention sublayers performs the worst, whereas the model with the 32-head self-attention sublayers performs the best, as far as the hyperparameters listed in Table~\ref{table_param} are concerned. We use both the model with the 1-head self-attention sublayers and the model with the 32-head self-attention sublayers in the following performance evaluation.

\begin{figure}[!t]
  \centering
  \includegraphics[width=0.7\columnwidth]{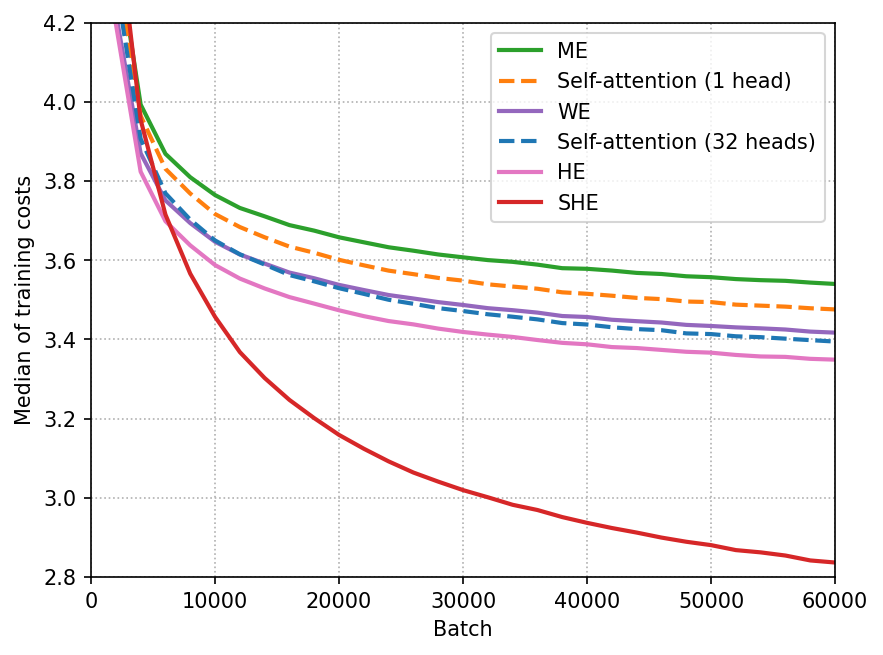} 
  \caption{Medians of the training costs of the models with either the self-attention sublayers or the proposed Extractor sublayers.}
  \label{fig_comp_medians}
\end{figure}
\begin{figure}[!t]
  \centering
  \includegraphics[width=0.7\columnwidth]{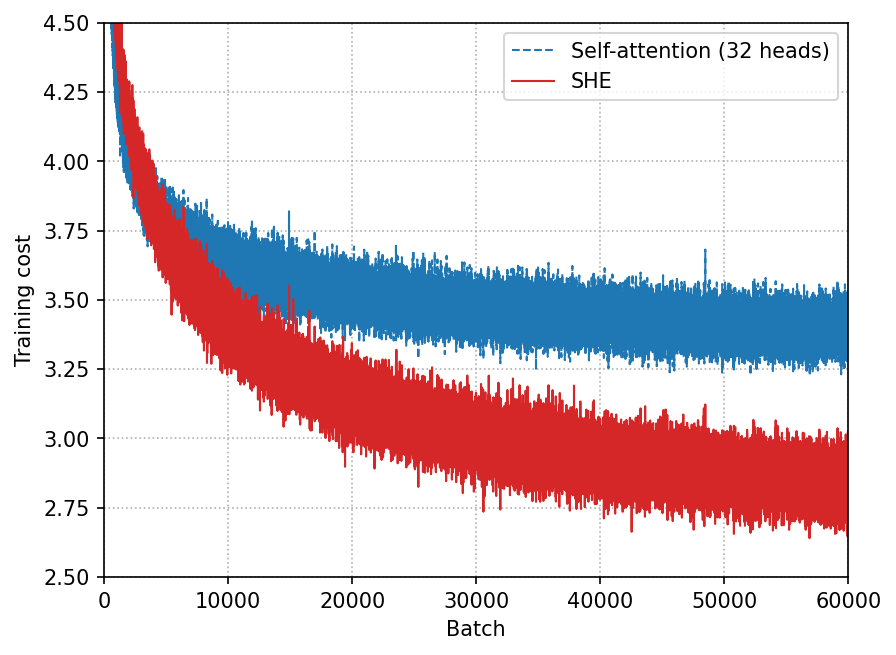} 
  \caption{Training costs of the models with either the 32-head self-attention sublayers or the proposed SHE sublayers.}
  \label{fig_comp_costs}
\end{figure}
Four Transformer models with the self-attention sublayers replaced by the proposed SHE/HE/WE/ME sublayers, respectively, are trained using the same hyperparameters and settings.  Fig.~\ref{fig_comp_medians} shows the medians of the training costs for every non-overlapping 2000 batches. Fig.~\ref{fig_comp_costs} shows the training costs of the models with either the 32-head self-attention sublayers or the proposed SHE sublayers for every batch. 
It can be observed that the proposed SHE evidently outperforms the self-attention mechanism. The simplified versions of the SHE, namely the HE/WE/ME, perform roughly as well as the self-attention mechanism. Specifically, in terms of the aforementioned hyperparameters and settings, the proposed HE outperforms the 32-head self-attention. Additionally, the performances of WE and ME are close to those of the 32-head self-attention and the 1-head self-attention, respectively.

For the purpose of human evaluation, we have the models with the SHE sublayers and the 32-head self-attention sublayers generate a sequence of text starting with the same given prompt ``Once upon a time there was a little princess who'' and using the same random seed. Top-$p$ sampling with $p$ set to 0.6 is employed. The generated texts are listed in Table 2. The quality of the generated texts generally improves as training cost drops.
\begin{table*}[t]
\centering
\begin{tabular}{l|l}
    \toprule
    Model & Generated text \\
   \midrule
    Transformer & Once upon a time there was a little princess who was so fond of a great feast that she always went \\
    ~~with the 32-  &  ~~to bed, and slept till morning. The next morning, when the emperor was asleep, he felt that it was    \\
    ~~head self-  &  ~~his duty to go to the emperor, and his eyes was full of pleasure. When he awoke he was in the  \\
    ~~attention  &  ~~habit of taking up his thoughts, and taking up his hands to listen to the idea of his own being  \\
    ~~sublayers &  ~~disposed of, and he gave up the news of the emperor's designs. He felt very much in love with ...  \\
   \midrule
   Transformer  & Once upon a time there was a little princess who was lying in a window. She was a little lonely   \\
   ~~with the &  ~~woman, and had found a little girl lying in a tree; but the old woman had never seen a golden  \\
   ~~proposed &  ~~apple, and she was too sleepy to find that the goose was so frightened that she could not see her  \\
   ~~SHE &  ~~eyes fall to the place where she could get it. `I am going to be a witch,' said she; `I will soon get    \\
   ~~sublayers &  ~~a princess and behave with all my other things.' So the old woman said to her that she was ... \\
   \bottomrule
\end{tabular}
\caption{Generated texts.}
\label{table_text}
\end{table*}

\begin{figure}[!t]
  \centering
  \includegraphics[width=0.7\columnwidth]{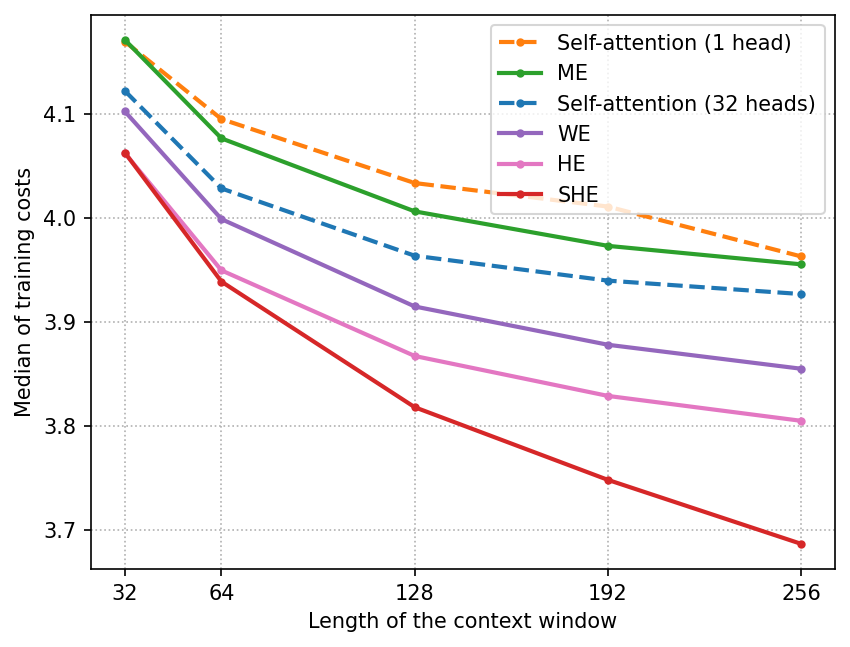} 
  \caption{Medians of the training costs of the models with different lengths of the context window (when $m=2$).}
  \label{fig_comp_length}
\end{figure}
\begin{figure}[!t]
  \centering
  \includegraphics[width=0.7\columnwidth]{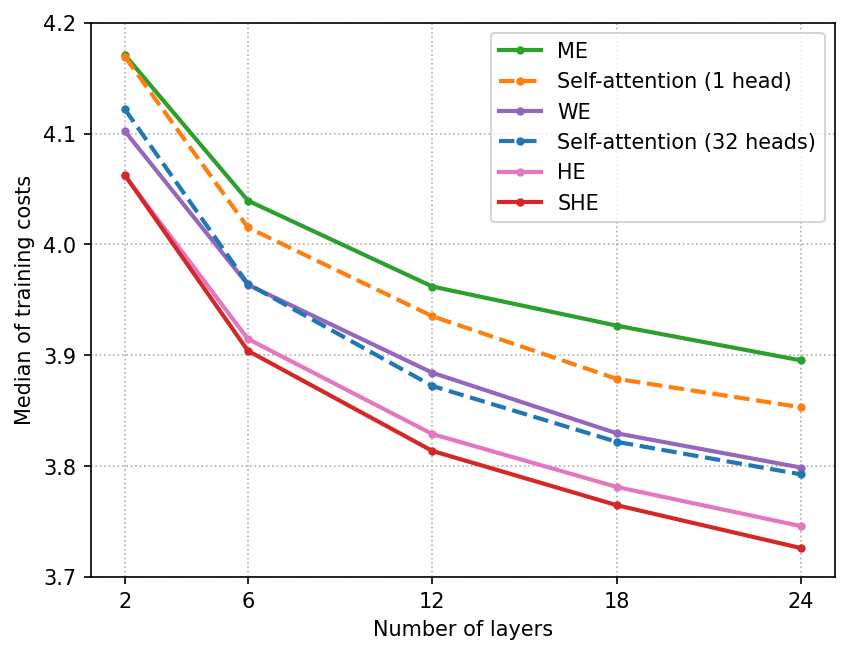} 
  \caption{Medians of the training costs of the models with different numbers of layers (when $l=32$).}
  \label{fig_comp_layer}
\end{figure}
Although the proposed SHE outperforms the self-attention mechanism under the hyperparameters and settings listed in Table~\ref{table_param}, we may wonder whether this conclusion still stands if other major hyperparameters such as the length of the context window and the number of layers are altered. In order to answer this question, models with different lengths of the context window and different numbers of layers are trained. The same hyperparameters and settings are used, except for the number of layers $m$ being reduced to 2, the length of the context window $l$ being reduced to 32, and the number of batches being reduced to 30000, for the sake of reducing the training time. Fig.~\ref{fig_comp_length} and Fig.~\ref{fig_comp_layer} show the medians of the training costs for the last 2000 batches. 
We can observe that, in general, the larger the length of the context window, the greater the performance gap between the models with the proposed SHE sublayers and the models with the 32-head self-attention sublayers. Additionally, both the models with the proposed Extractor sublayers and the models with the self-attention sublayers perform better as the number of layers increases. Both figures indicate the effectiveness of the proposed Extractors.

\section{Computational Complexity}

Both performance and computational complexity matter. In this section, we estimate the computational complexities of the proposed Extractor sublayers and compare them with that of the self-attention sublayer. 

\begin{table*}[t]
\centering
\begin{tabular}{l|l|l}
    \toprule
    Number of & The self-attention sublayer & The proposed SHE sublayer  \\ 
   \midrule
    Multiplications  & $(3d\cdot\frac{d}{n}\cdot l+\sum_{t=1}^{l}(t\cdot\frac{d}{n}+t\cdot\frac{d}{n}))\cdot n+d\cdot d\cdot l$ & $\sum_{t=1}^{l}{(d\cdot d\cdot t+d\cdot d+d+d\cdot d)}$  \\ 
     & ~~$=l^2d+4ld^2+ld$ & ~~$=\frac{1}{2}l^2d^2+\frac{5}{2}ld^2+ld$  \\ 
   \midrule
   Additions & $(\sum_{t=1}^{l}(t\cdot(\frac{d}{n}-1)+(t-1)\cdot\frac{d}{n}+(t-1))$ & $\sum_{t=1}^{l}((d-1)\cdot d\cdot t+(t-1)\cdot d$  \\
    & ~~$+3(d-1)\cdot\frac{d}{n}\cdot l )\cdot n+(d-1)\cdot d\cdot l$ & ~~$+(d-1)\cdot d+(d-1)\cdot d)$  \\
    & ~~$=l^2d+4ld^2-4ld-ln$ & ~~$=\frac{1}{2}l^2d^2+\frac{5}{2}ld^2-3ld$ \\
   \midrule
   Divisions & $n\sum_{t=1}^{l}\left(t+t\right)=nl^2+nl$ &  $0$  \\
   \midrule
   Exponentiations & $n\sum_{t=1}^{l}t=\frac{1}{2}nl^2+\frac{1}{2}nl$ & $0$  \\
   \midrule
   Parameters  & $3d\cdot \frac{d}{n}\cdot n+d\cdot d=4d^2$ & $d\cdot d\cdot l+d\cdot d+d\cdot d=ld^2+2d^2$  \\
   \bottomrule
\end{tabular}
\caption{Computational complexity in training (part I).}
\label{table_train_1}
\end{table*}
\begin{table*}[t]
\centering
\begin{tabular}{l|l|l|l}
   \toprule
    Number of & The proposed HE sublayer & The proposed WE sublayer & The proposed ME sublayer \\ 
   \midrule
    Multiplications  &    $\sum_{t=1}^{l}{(d\cdot t+d\cdot d+d+d\cdot d)}$    &    $\sum_{t=1}^{l}{(d\cdot t+d\cdot d+d+d\cdot d)}$    &    $\sum_{t=1}^{l}{(d\cdot t)}$    \\ 
     & ~~$+d\cdot d\cdot l$    &    ~~$=\frac{1}{2}l^2d+\frac{3}{2}ld+2ld^2$    &    ~~$=\frac{1}{2}l^2d+\frac{1}{2}ld$  \\
     & ~~$=\frac{1}{2}l^2d+\frac{3}{2}ld+3ld^2$    &        &      \\ 
   \midrule
   Additions &    $\sum_{t=1}^{l}{((t-1)\cdot d+(d-1)\cdot d}$    &    $\sum_{t=1}^{l}{((t-1)\cdot d+(d-1)\cdot d}$    &    $\sum_{t=1}^{l}{((t-1)\cdot d)}$ \\
    & ~~$+(d-1)\cdot d)+(d-1)\cdot d\cdot l$    &    ~~$+(d-1)\cdot d)$    &    ~~$=\frac{1}{2}l^2d-\frac{1}{2}ld$ \\
    & ~~$=3ld^2+\frac{1}{2}l^2d-\frac{7}{2}ld$    &    ~~$=2ld^2+\frac{1}{2}l^2d-\frac{5}{2}ld$    &    \\
   \midrule
   Divisions & $0$ & $0$ & $0$ \\
   \midrule
   Exponentiations & $0$ & $0$ & $0$ \\
   \midrule
   Parameters  & $d\cdot l+d\cdot d+d\cdot d+d\cdot d$ & $d\cdot l+d\cdot d+d\cdot d$  & $l$ \\
    & ~~$=ld+3d^2$ & ~~$=ld+2d^2$ & \\
   \bottomrule
\end{tabular}
\caption{Computational complexity in training (part II).}
\label{table_train_2}
\end{table*}According to the architectures illustrated in Fig.~\ref{fig_att}, Fig.~\ref{fig_ext_xl}, Fig.~\ref{fig_ext_l}, Fig.~\ref{fig_ext_m}, and Fig.~\ref{fig_ext_s}, given an input sequence of length $l$ during training, the numbers of multiplications, additions, divisions, exponentiations, and trainable parameters of the Extractor sublayers and the self-attention sublayer are estimated in Table~\ref{table_train_1} and Table~\ref{table_train_2}. 
During inference,  given a new input token at sequence length $t$, the incremental numbers of multiplications, additions, divisions, and exponentiations of the Extractor sublayers and the self-attention sublayer are estimated in Table~\ref{table_infer_1} and Table~\ref{table_infer_2}.
\begin{table*}[t]
\centering
\begin{tabular}{l|l|l}
    \toprule
    Number of & The self-attention sublayer & The proposed SHE sublayer  \\ 
   \midrule
    Multiplications  & $\left(3d\cdot\frac{d}{n}+t\cdot\frac{d}{n}+t\cdot\frac{d}{n}\right)\cdot n+d\cdot d$ & $d\cdot d\cdot t+d\cdot d+d+d\cdot d$  \\ 
     &  ~~$=2td+4d^2$ & ~~$=td^2+2d^2+d$  \\ 
   \midrule
   Additions & $(3(d-1)\cdot\frac{d}{n}+t\cdot(\frac{d}{n}-1)+(t-1)\cdot\frac{d}{n}$ & $(d-1)\cdot d\cdot t+(t-1)\cdot d+(d-1)\cdot d$  \\
    & ~~$+(t-1))\cdot n+(d-1)\cdot d$ & ~~$+(d-1)\cdot d$  \\
    & ~~$=2td-tn+t+4d^2-5d-1$ & ~~$=td^2+2d^2-3d$  \\
   \midrule
   Divisions & $(t+t)\cdot n=2tn$ & $0$  \\
   \midrule
   Exponentiations & $t\cdot n=tn$ & $0$  \\
   \bottomrule
\end{tabular}
\caption{Computational complexity in inference (part I).}
\label{table_infer_1}
\end{table*}
\begin{table*}[t]
\centering
\begin{tabular}{l|l|l|l}
    \toprule
    Number of &  The proposed HE sublayer & The proposed WE sublayer & The proposed ME sublayer \\ 
   \midrule
    Multiplications  & $d\cdot t+d\cdot d+d\cdot d+d$ & $d\cdot t+d\cdot d+d+d\cdot d$ & $d\cdot t=td$ \\ 
     &  ~~$+d\cdot d$ & ~~$=td+2d^2+d$  & \\ 
     &  ~~$=td+3d^2+d$ &  & \\ 
   \midrule
   Additions & $(t-1)\cdot d+(d-1)\cdot d$ & $(t-1)\cdot d+(d-1)\cdot d$ & $(t-1)\cdot d=td-d$  \\
    & ~~$+(d-1)\cdot d+(d-1)\cdot d$ & ~~$+(d-1)\cdot d$  & \\
    & ~~$=td+3d^2-4d$ & ~~$=td+2d^2-3d$  & \\
   \midrule
   Divisions & $0$ & $0$ & $0$  \\
   \midrule
   Exponentiations & $0$ & $0$ & $0$ \\
   \bottomrule
\end{tabular}
\caption{Computational complexity in inference (part II).}
\label{table_infer_2}
\end{table*}

\begin{table*}[t]
\centering
\begin{tabular}{l|l|l|l|l|l|l}
   \toprule
    Number of & The 1-head & The 32-head   & The SHE  & The HE & The WE & The ME \\ 
     & self-attention & self-attention & ~~sublayer  & ~~sublayer & ~~sublayer & ~~sublayer \\ 
     & ~~sublayer & ~~sublayer   &    &    &    &     \\
   \midrule
    Multiplications  &    $10,502,144$  &  $10,502,144$  &  $139,476,992$  & $7,364,608$  &    $5,267,456$    &    $1,056,768$    \\ 
   \midrule
   Additions &    $10,420,096$   &   $10,416,128$   & $139,411,456$ & $7,282,688$  &    $5,201,920$    &    $1,040,384$ \\
   \midrule
   Divisions & $16,512$ & $528,384$ & $0$ & $0$ & $0$  & $0$  \\
   \midrule
   Exponentiations & $8,256$ & $264,192$ & $0$ & $0$ & $0$  & $0$  \\
   \midrule
   Parameters  & $65,536$ & $65,536$ & $2,129,920$  & $65,536$  &  $49,152$  &  $128$ \\
   \midrule
   Total arithmetic & $20,947,008$ & $21,710,848$ & $278,888,448$  & $14,647,296$  &  $10,469,376$ & $2,097,152$ \\
   ~~operations  &  &  &   &   &   &  \\
   \bottomrule
\end{tabular}
\caption{Number of trainable parameters and total arithmetic operations in training when $d=128$ and $l=128$.}
\label{table_train_number}
\end{table*}
When substituting $d=128$ and $l=128$, which is the case shown in Table~\ref{table_param}, into Table~\ref{table_train_1} and Table~\ref{table_train_2}, we obtain the numbers of total arithmetic operations and trainable parameters, as listed in Table~\ref{table_train_number}. We can see that both the number of total arithmetic operations and the number of trainable parameters of the proposed SHE sublayer are much larger than those of the self-attention sublayer. This explains why the proposed SHE is capable of outperforming the self-attention mechanism, as the saying goes, ``you get what you pay for''. The proposed HE outperforms the 32-head self-attention mechanism with fewer arithmetic operations and the same number of trainable parameters. This suggests that the proposed Extractors are indeed capable of outperforming the self-attention mechanism. Additionally, the proposed WE outperforms the 1-head self-attention mechanism with fewer arithmetic operations and trainable parameters. The performance of the proposed ME is close to that of the 1-head self-attention mechanism, with much fewer arithmetic operations and trainable parameters (approximately $\frac{1}{10}$ and $\frac{1}{512}$, respectively).

In practice, given sufficient computing power, the critical path of computation matters. Take the inference phase as an example. According to Fig.~\ref{fig_att}, the critical path of the self-attention sublayer is ``multiplication - cumulation - multiplication - cumulation - division - exponentiation - cumulation - division - multiplication - cumulation - multiplication - cumulation'', whereas  according to Fig.~\ref{fig_ext_xl} the critical path of the SHE sublayer is ``multiplication - cumulation - multiplication - multiplication - cumulation''. The critical path of the SHE sublayer is shorter than that of the self-attention sublayer, meaning the proposed SHE sublayer has the potential to run faster than the self-attention sublayer. Moreover, according to Fig.~\ref{fig_ext_s} the critical path of the ME sublayer is ``multiplication - cumulation'', which is much shorter.

\section{Conclusion}

In this paper, a family of drop-in replacements for the existing self-attention mechanism in the Transformer, called the Extractors, is proposed and evaluated. Specifically, four versions of the Extractors, namely the SHE, the HE, the WE, and the ME, are proposed as examples. 

Experimental results show that the proposed SHE evidently outperforms the self-attention mechanism. Although the SHE requires more trainable parameters and computations, it has a shorter critical path of computation and thus has the potential to run faster, providing a way to significantly boost the performance of the Transformer. The proposed HE and WE are capable of outperforming the multi-head self-attention and the 1-head self-attention, respectively, with fewer arithmetic operations and trainable parameters. They (as well as their variants) are ideal candidates for replacing the self-attention mechanism. The proposed ME is suitable for computation-constrained scenarios since it requires much fewer arithmetic operations and trainable parameters while maintaining a performance close to that of the 1-head self-attention.

Furthermore, the sequence prediction problem in the context of text generation is formulated using variable-length discrete-time Markov chains, and the Transformer is reviewed based on our understanding.

We hope this work will contribute to building powerful and cost-effective Transformer-based large models.
Moreover, it is anticipated that more research will be carried out to improve the performance of the Transformer or address the sequence prediction problem inspired by this work.

\bibliographystyle{unsrtnat}
\bibliography{references}  

\end{document}